\documentclass[letterpaper, 10 pt, journal, twoside]{IEEEtran}

\usepackage{cite}
\usepackage{amsmath,amssymb,amsfonts}
\usepackage{algorithm}
\usepackage{algpseudocode}
\usepackage{graphicx}
\usepackage{textcomp}
\usepackage{xcolor}
\usepackage{amsmath} 
\usepackage{amssymb}  
\usepackage{comment}
\usepackage{siunitx}
\usepackage{subcaption}
\usepackage{caption}
\usepackage{multirow}
\usepackage{hyperref}
\usepackage{booktabs}
\usepackage{makecell}
\usepackage{soul}
\usepackage{threeparttable}
\usepackage{bm}
\usepackage{arydshln} 

\usepackage[capitalize]{cleveref}
\crefname{section}{Sec.}{Secs.}
\Crefname{section}{Section}{Sections}
\Crefname{table}{Table}{Tables}
\crefname{table}{Tab.}{Tabs.}

\IEEEoverridecommandlockouts                              

\makeatletter
\usepackage{xspace}
\DeclareRobustCommand\onedot{\futurelet\@let@token\@onedot}
\def\@onedot{\ifx\@let@token.\else.\null\fi\xspace}

\def\etal{{et al}\onedot}
\makeatother
\def\etalcite#1{\etal~\cite{#1}}

\algdef{SE}[FOR]{ForEach}{EndForEach}[1]{\textbf{foreach} #1\ \algorithmicdo}{\algorithmicend}%

\title{GOReloc: Graph-based Object-Level Relocalization for Visual SLAM} 

\author{Yutong Wang$^{1}$, Chaoyang Jiang$^{1}$, Xieyuanli Chen$^{2}$
\thanks{Manuscript received: June 3, 2024; Accepted: August 1, 2024.}
\thanks{This paper was recommended for publication by Editor C. Javier upon evaluation of the Associate Editor and Reviewers’ comments.} \thanks{This work was supported by the National Key Research and Development Project of China (No. 2020YFC1512503) and the National Natural Science Foundation of China (No. U20A20333). (Corresponding authors: Chaoyang Jiang and Xieyuanli Chen.)}
\thanks{$^{1}$Y. Wang and C. Jiang are with the School of Mechanical Engineering, Beijing Institute of Technology, China, and the Yangtze Delta Region Academy of Beijing Institute of Technology, Jiaxing, China.
        {\tt\small yutongwang@bit.edu.cn, cjiang@bit.edu.cn}}%
\thanks{$^{2}$X. Chen is with the College of Intelligence Science and Technology, National University of Defense Technology, China
        {\tt\small xieyuanli.chen@nudt.edu.cn}}%
\thanks{Digital Object Identifier (DOI): 10.1109/LRA.2024.3442560.}
}

\begin{document}
\maketitle

\markboth{IEEE ROBOTICS AND AUTOMATION LETTERS. PREPRINT VERSION. ACCEPTED August, 2024}{Wang \MakeLowercase{\textit{et al.}}: GOReloc: Graph-based Object-Level Relocalization for Visual SLAM}

\begin{abstract}

This article introduces a novel method for object-level relocalization of robotic systems. It determines the pose of a camera sensor by robustly associating the object detections in the current frame with 3D objects in a lightweight object-level map. Object graphs, considering semantic uncertainties, are constructed for both the incoming camera frame and the pre-built map. Objects are represented as graph nodes, and each node employs unique semantic descriptors based on our devised graph kernels. We extract a subgraph from the target map graph by identifying potential object associations for each object detection, then refine these associations and pose estimations using a RANSAC-inspired strategy. Experiments on various datasets demonstrate that our method achieves more accurate data association and significantly increases relocalization success rates compared to baseline methods. The implementation of our method is released at \url{https://github.com/yutongwangBIT/GOReloc}.

\end{abstract}

\begin{IEEEkeywords}
SLAM, Localization, Object-Level SLAM, Relocalization.
\end{IEEEkeywords}
\section{Introduction}
\IEEEPARstart{V}{isual} simultaneous localization and mapping~(SLAM) enables robots to understand and navigate the environments accurately and efficiently~\cite{mur2017orb, engel2017direct}. Traditional geometric SLAM approaches, which rely on features or optical flow, often overlook the contextual and semantic understanding of the environment. To address this limitation, semantic SLAM has been developed by incorporating semantic information into the process, enhancing interaction with diverse environmental elements~\cite{salas2013slam, chen2019iros}. Within semantic SLAM research, there is a growing preference for object-oriented SLAM over dense voxel-based approaches, as the former offers more lightweight and computationally efficient representations. Typical object-oriented systems, such as CubeSLAM~\cite{yang2019cubeslam} and QuadricSLAM~\cite{nicholson2018quadricslam}, adopt geometric simplifications like cubes or quadrics to describe the detected objects, leading to a significant reduction in computational resources.

\begin{figure}[t]
	\centering
 \includegraphics[width=0.99\linewidth]{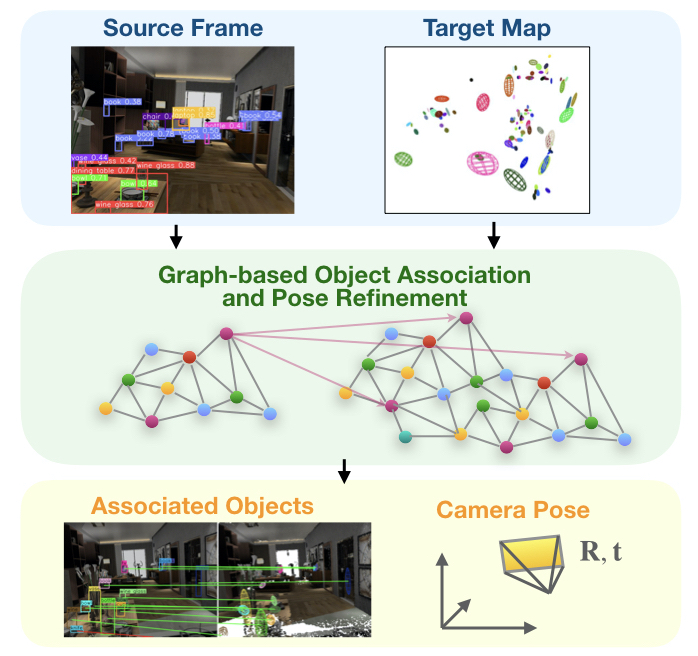}
	\caption{Our system relocalizes a camera frame in a lightweight, object-level map via graph-based approaches.}
	\label{fig:motivation}
 \vspace{-0.2cm}
\end{figure}

Incorporating object information can not only enhance the robustness of odometry and local mapping in SLAM but also potentially improve the success rate of relocalization. The emphasis on relocalization, which is the process where a mobile platform determines its position using a pre-built map, plays a crucial role in advancing long-term navigation in robotic systems~\cite{williams2011automatic}. Traditional relocalization methods relying on feature points often face challenges in adapting to variations in illumination and changes in viewpoint~\cite{galvez2012bags}. Incorporating advanced object information into relocalization processes significantly enhances their robustness, highlighting the importance and value of research in object-level relocalization.

In the realm of visual object-level relocalization, traditional non-graph-based methods struggle with data association, often relying on brute-force combinations and geometric validation to identify the best solutions~\cite{zins2022oa,Mahattansin2022Reloc}. These methods can be inefficient and less effective, particularly in complex or dynamically changing environments. In contrast, graph-based approaches~\cite{gawel2018x,liu2019global,yu2022semanticloop,kabalar2023towards,zhao2023graph}
enhance robustness by using topological information for object matching, which refines pose estimation through alignment. However, common graph matching techniques like random walk and Weisfeiler-Lehman graph kernel are error-prone, especially in environments with repetitive scenes~\cite{stumm2016robust}, impacting the overall performance of object-level relocalization.

To address the abovementioned issues in object-level relocalization and enhance its practicality in real-world scenarios, we present GOReloc, a Graph-based Object-level Relocalization system for visual SLAM. Our method effectively relocalizes by robustly associating objects between the RGB image of a camera frame and the object-level map using dual quadric models~\cite{wang2023qiso}, as shown in~\cref{fig:motivation}. GOReloc innovatively incorporates semantic uncertainty and consistency into the graph construction and kernel computation processes. This method can increase the distinctiveness of node descriptors, thereby enhancing node-matching accuracy. Moreover, rather than relying on pairwise matching, our approach extracts a subgraph from the target map graph by identifying a series of potential objects for each detection of the source frame graph. Finally, a RANSAC-inspired method is proposed to refine pose estimation and object associations, iteratively selecting subsets of detections and candidates to ensure robust performance. Our key contributions are threefold:
\begin{itemize}
    \item Integrated semantic uncertainty and consistency into graph generation and graph kernel computation, enhancing the distinctiveness of node descriptors and facilitating more precise object matching.
    \item Explored multiple potential matches to extract subgraphs, followed by a novel refinement strategy to determine the best relocalization results, reducing its reliance on direct graph matching.
    \item Built a complete system capable of real-time processing, ensuring efficiency for practical applications. Comprehensive experimental results validate its effectiveness.
\end{itemize}
\section{Related Work}
Object-level relocalization methods in visual SLAM systems are broadly classified into non-graph and graph-based approaches. Among the non-graph-based ones, Li~\etalcite{li2019semantic} employed the Hungarian algorithm for optimal matching of object landmarks between two semantic maps, considering label differences and spatial proximity. Mahattansin~\etalcite{Mahattansin2022Reloc} introduced a high-level feature array consisting of object classes and employed it as a constraint to identify relocalization candidates. OA-SLAM by Zins~\etalcite{zins2022oa} proposed a complete object-aided relocalization approach that identifies the camera frame pose in an object-level map. It iterates all feasible detection-object match pairs and finds the best results with the lowest projection error. However, such non-graph-based methods substantially increase computational load as the scene size and number of objects grow, hindering its practical application in real-world scenarios.

Graph-based methods for object-level relocalization have been proven to enhance robustness. The concept of a graph involves a set of nodes connected by edges, which is used to represent relationships or connections between different entities. In the context of lightweight, object-level relocalization, graph-based approaches excel by integrating diverse data in their nodes --- not just 1D labels and 2D detections, but also 3D geometric models. Meanwhile, the edges represent the spatial relationships between objects. Some methods~\cite{liu2019global,yu2022semanticloop} modeled the objects as dense semantic features, such as TSDF volumes, and the query image or submap and the global map are represented as graphs with semantics and topology. Then, the methods achieve global localization through precise object alignment. In work by Lin~\etalcite{lin2021topology}, rather than employing dense semantic models, a lightweight cuboid model is utilized for object representation and graph extraction. Its loop detection employs edit distance to match the random walk descriptor vectors, and loop correction involves aligning cuboid objects across semantic maps. Wu~\etalcite{wu2023object} also employ a random-walk descriptor that integrates semantic labels, object size, distance, and angle to construct a unified descriptor vector. 

However, such graph-based approaches are often sensitive to noise and outliers, as semantic labels can be uncertain and inconsistent. Moreover, most of the graph-based methods directly estimate camera poses from graph-matching results, which leads to significant pose errors when mismatches occur due to semantic and topological similarities. In this paper, we firstly enhance the robustness of graph matching by integrating semantic uncertainty and consistency in graphs. Furthermore, we also introduce a subgraph extraction strategy that retains multiple object candidates per detection, followed by a refinement process for both object associations and camera pose to improve relocalization accuracy. 
\section{Our Approach: GOReloc}
\label{sec_system}

\begin{figure*}[t]
	\centering
		\includegraphics[width=1\linewidth]{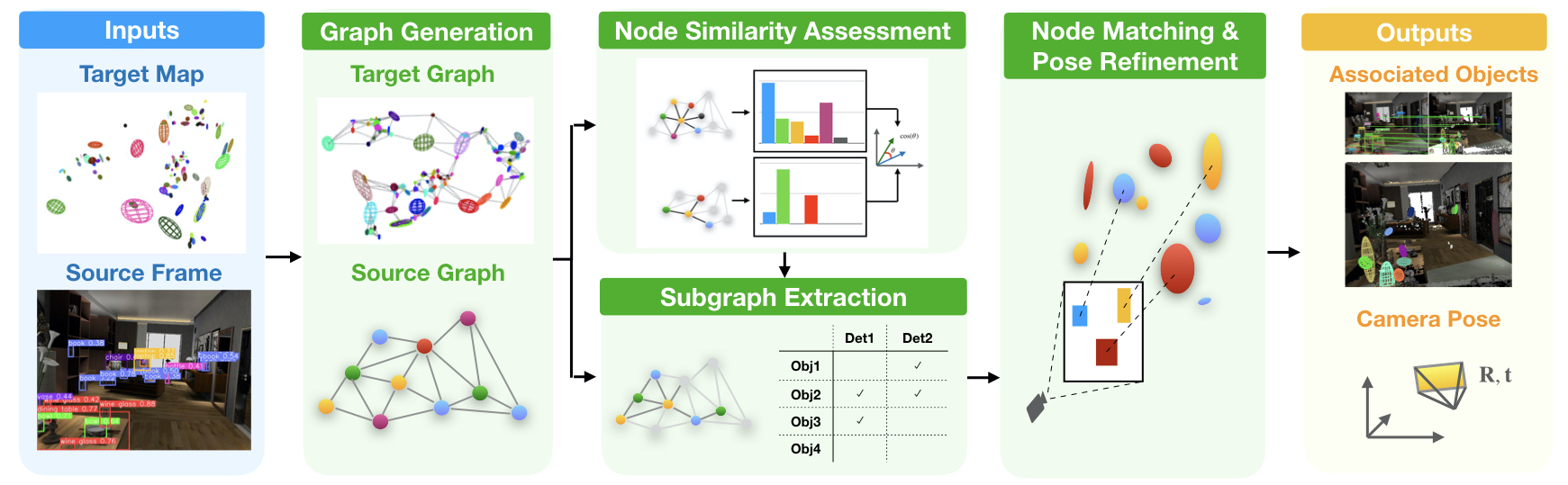}
	\caption{Workflow of our proposed GOReloc. The system relocalizes a camera frame with object detections in an object-level map. After generating the object graphs considering semantic uncertainty and consistency, a subgraph from the target map graph is extracted by selecting association candidates based on node similarity assessment. The final of pose estimation and object associations are refined using a RANSAC-inspired method.}
	\label{fig:pipeline}
    \vspace{-0.2cm}
\end{figure*}

\subsection{Problem Definition}
Our main focus in object-level relocalization is to estimate the camera pose of a source frame in a target object map using graph-based approaches, relying solely on the RGB image of the source frame. This task hinges on associating current frame detections with 3D objects in the map, which contain both geometric and semantic information. 

Let $\mathcal{F} = \{\mathcal{D}, \mathcal{G}^s\}$ represent the input source frame and $\mathcal{M} = \{\mathcal{O}, \mathcal{K}, \mathcal{G}^t\}$ the target 3D object map. In this context, $\mathcal{D}=\{d_j\}^N_{j=1}$ denotes the object detection results provided by YOLOv8~\cite{Redmon2016YOLO}, and $\mathcal{G}^s$ represents the source frame graph. The object-level map is constructed using the mapping method proposed in VOOM~\cite{wang2024icra}, where the set of map objects $\mathcal{O}=\{o_i\}^M_{i=1}$ represented by dual quadrics and the keyframes $\mathcal{K}$ that observe the objects are stored. Each object $o_i \in \mathcal{O}$ contains both semantic information and geometric parameters $\vec{q}_i = [\vec{t}_i, \vec{\theta}_i, \vec{s}_i]$, where $\vec{t}_i$, $\vec{\theta}_i$ and $\vec{s}_i$ are the position, orientation and size vector, respectively. Additionally, $\mathcal{G}^t$ represents the target map graph. 

In this paper, we define the object-level relocalization challenge as a dual task of the object-level association $\mathcal{A}$ and the camera pose estimation $\mathbf{T}$ based on the information of $\mathcal{F}$ and $\mathcal{M}$. Specifically, the object-level association $\mathcal{A} = \{(n^s_j, n^t_i)\}$ represents the matching object pairs between source frame graph $\mathcal{G}^s$ and target map graph $\mathcal{G}^t$. Camera pose estimation, represented by the transformation matrix $\mathbf{T}$, is derived by estimating the transformation from the source frame to the target map.

\subsection{System Overview}
Our GOReloc system, depicted in~\cref{fig:pipeline}, starts by taking an object-level map and a source frame as inputs. The initial step in the proposed GOReloc is to establish a target map graph $\mathcal{G}^t$ and a source frame graph $\mathcal{G}^s$ taking into account semantic uncertainties, detailed in~\cref{subsec_gen}. In such graphs, Nodes represent objects or detections, and edges denote their topological relationships. Once these graphs are created, the system focuses on extracting a subgraph from the target map graph. This is accomplished by selecting association candidates through node similarity assessment, which identifies the most relevant objects for comparison based on graph kernels. Details for node similarity assessment via graph kernels are presented in~\cref{subsec_similarity}, while the subgraph extraction method is introduced in ~\cref{subsec_subgraph}. To further enhance accuracy, the system employs a RANSAC-inspired method to refine both the pose estimation and the object associations, as described in ~\cref{subsec_refine}. This method iteratively selects subsets of the detections and candidates to estimate the optimal camera pose and object associations, effectively filtering out outliers and ensuring robust performance.

\subsection{Graph Generation with Semantics}
\label{subsec_gen}
\subsubsection{Source Frame Graph Generation  Considering Semantic Uncertainty}
The graph $\mathcal{G}^s$ for the source frame, constructed based on the filtered object detection results $\mathcal{D}$ from YOLOv8, comprises node and edge information. 
We filter object detection results by keeping only those with confidence above $0.1$ and removing bounding boxes with overlapping areas greater than $0.6$, retaining the one with the higher confidence. Each node $n_j$ in the graph represents the $j$-th detection $d_j$, consisting of a bounding box $\text{bbox}_j$, a label $l_j$, and a confidence score $s_j$. Initially, an inscribed ellipse $\text{ell}_j$ of $\text{bbox}_j$ is calculated. We then define a probability $P(c | n_j)$ to denote the likelihood of node $n_j$ belonging to a category $c$ within the whole semantic category set $\mathcal{C}$: 
\begin{equation}
\label{eqa_prob_base}
P(c | n_j) = s_j \cdot \delta(c, l_j),
\end{equation}
where $\delta(c, l_j)$ is an indicator function that equals $1$ if $l_j$ matches $c$, and $0$ otherwise. Consequently, such information $\mathcal{I}_j = (\text{ell}_j, l_j, \{P(c|n_j) \}_{c\in \mathcal{C}})$ is stored for the node $n_j$. When the nodes are generated, edges are then created by connecting each node to its $K$ nearest neighboring nodes. The distance between nodes is measured as the Euclidean distance between the centers of their bounding boxes, and this distance serves as the edge weight $w$.

\subsubsection{Target Map Graph Generation Considering Semantic Consistency}
The target map comprises keyframes $\mathcal{K}$ and objects $\mathcal{O}$, which are geometrically represented by dual quadrics. To construct a target map graph $\mathcal{G}^t$, the semantic information for each object $o_i$ is also necessary. As noted in VOOM~\cite{wang2024icra}, an object may be assigned with different labels due to the uncertainty of object detection. Given that the graph is constructed based on these categories, it is crucial to account for this uncertainty considering multi-labels, particularly in large-scale scenarios. Ensuring semantic consistency across observation keyframes can help mitigate the impact of such uncertainties. First, each object $o_i$ is corresponded with a node $n_i$ in $\mathcal{G}^t$. For each object $o_i$ observed within the set of keyframe indices $\mathcal{K}_i$, all associated detections $\{d_{ik}\}_{k \in \mathcal{K}_i}$ can be recorded in the map via VOOM. Each $d_{ik}$ has a corresponding label $l_{ik}$ and probability set $\{P(c | d_{ik})\}_{c\in\mathcal{C}}$, where each $P(c | d_{ik})$ is computed as in~\cref{eqa_prob_base}. Then, the likelihood $P(c | n_i)$ that node $n_i$ corresponding to object $o_i$ is associated with an arbitrary category $c$ can be determined through statistical analysis:
\begin{equation}
\label{eqa_prob_multi}
    P(c | n_i) = \frac{\sum_{k \in \mathcal{K}_i} P(c | d_{ik})}{\sum_{k \in \mathcal{K}_i} \sum_{c \in \mathcal{C}} P(c | d_{ik})}.
\end{equation}
Subsequently, we can summarize both geometric and semantic information for each object node in the target map as $\mathcal{I}_i = (\vec{q}_i, \{P(c | n_i)\}_{c \in \mathcal{C}})$. We then establish edges by connecting each object to its $K$ nearest neighboring objects. The edge weights $w$ are defined based on the Euclidean distances between each node pair.

\begin{figure*}
	\centering
 \begin{subfigure}[t]{0.38\textwidth}
    \centering	\includegraphics[height=3.8cm]{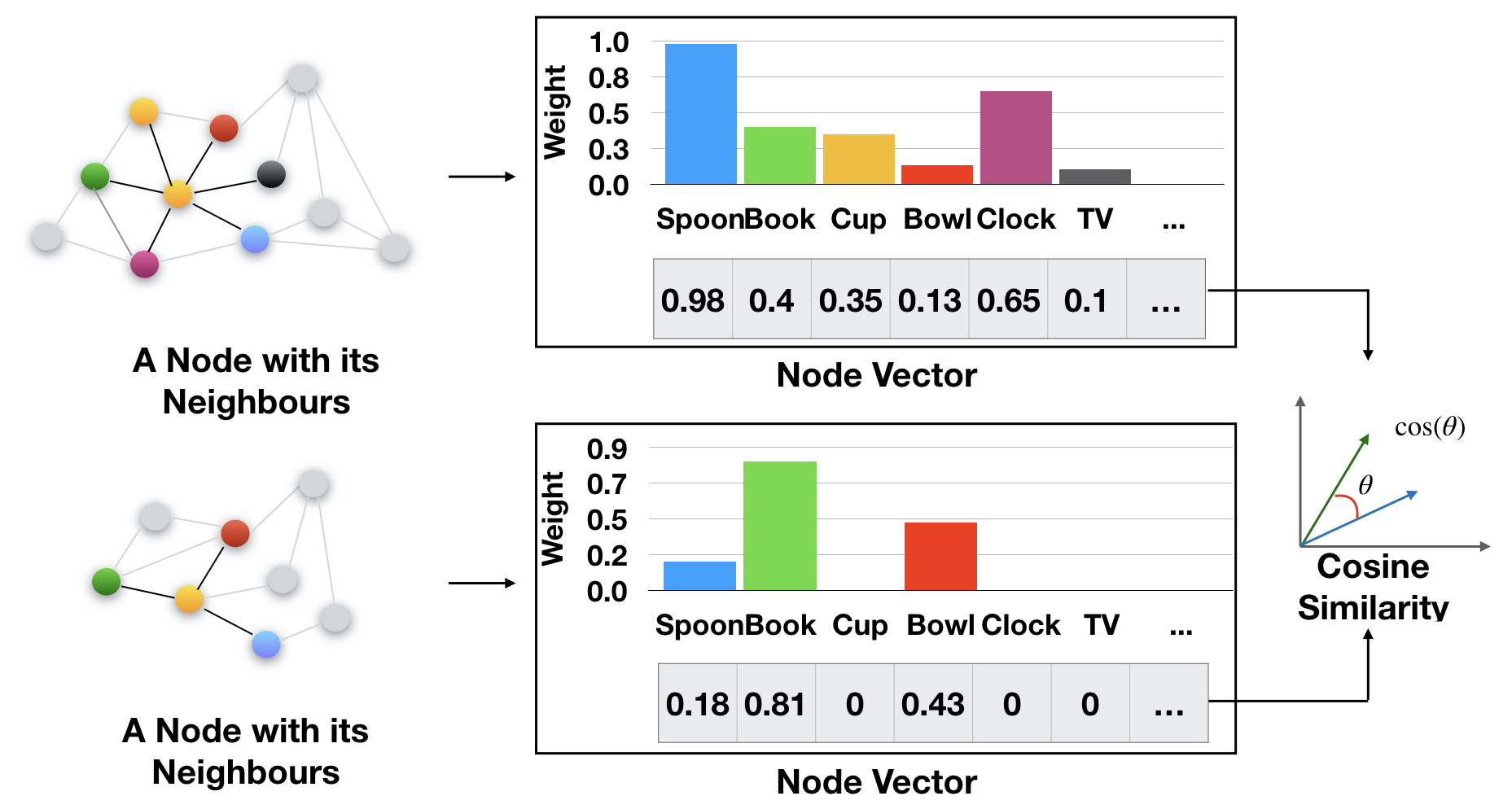}
	\caption{Node similarity calculation.}
		\label{fig:node_sim}
	\end{subfigure} \noindent \hspace{1pt}
	\begin{subfigure}[t]{0.6\textwidth}
	\centering	
        \includegraphics[height=3.8cm]{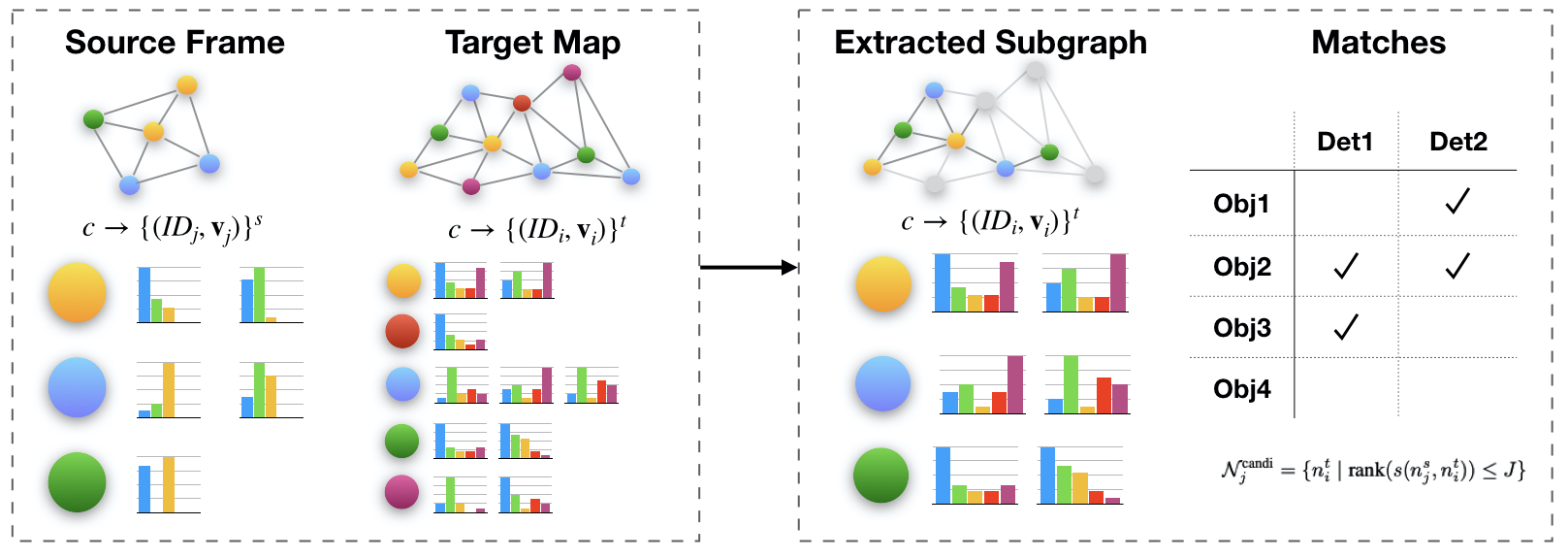}		
	\caption{Subgraph Extraction.}
		\label{fig:selection}
	\end{subfigure}
    \vspace{-0.05cm}
	\caption{Illustrations of node similarity assessment via graph kernels and subgraph extraction. The colors of the nodes in the graph distinguish between different categories of detections or objects.}    
   \vspace{-0.2cm}
\end{figure*}
\subsection{Node Similarity Assessment via Graph Kernels}
\label{subsec_similarity}
The design of our proposed node similarity calculation via graph kernels is illustrated in~\cref{fig:node_sim}. At first, the graph kernel construction is motivated by~\cite{stumm2016robust}. For both the source frame graph $\mathcal{G}^s$ and the target map graph $\mathcal{G}^t$, each node can serve as a root node $n_r$, and we relabel the node with a vector $\vec{v}$ based on the combination of its categories of its neighbors $\mathcal{N}_r$. The dimension of each vector is equal to the total number of semantic categories $\mathcal{C}$. The $k$-th value $\vec{v}_k$ in the vector, corresponding to the $k$-th category $c_k$, is accumulated based on the edge weights between the root node and its neighbors. To be more specific, the formula for calculating $\vec{v}_k$ is as follows:
\begin{equation}
\label{eq_vec}
    \vec{v}_k = \sum_{n_i\in \mathcal{N}_r}{w(n_r, n_i)P(c_k | n_i)},
\end{equation}
where $\mathcal{N}_r$ denotes the neighbor nodes set of the root node $n_r$, and $w(n_r, n_i)$ represents the edge weight between the root node and one of its neighbor nodes $n_i$. And $P(c_k | n_i)$ is calculated with~\eqref{eqa_prob_base} and~\eqref{eqa_prob_multi} for their respective source frame graph and target map graph. This function has employed both the edge weight $w(n_r, n_i)$ and the likelihood $P(c_k | n_i)$, representing the semantic uncertainty and consistency of the predicted label, to weigh the graph kernel descriptors. This strategy enhances descriptor distinctiveness, thereby improving the robustness of the subsequent node comparison and pose refinement process. 

Then, it is also essential to normalize the vector $\vec{v}$ after traversing all the neighbors of the root node to ensure the reliability and consistency of node comparisons. When calculating the similarity score between a node $n^s_j$ in $\mathcal{G}^s$ and a node $n^t_i$ in $\mathcal{G}^t$, we use the cosine distance between their respective normalized vectors, which is given by:
\begin{equation}
\label{eq_sim}
    s(n^s_j, n^t_i) = 1 - \frac{\vec{v}_{n^s_j} \cdot \vec{v}_{n^t_i}}{\|\vec{v}_{n^s_j}\| \|\vec{v}_{n^t_i}\|}
\end{equation}
\subsection{Subgraph Extraction from Target Map}
\label{subsec_subgraph}
The target map can be extensive in real-world scenarios and may contain similar patterns. Directly matching a small source frame graph with the target map graph can be inefficient and inaccurate. Therefore, subgraph extraction is crucial for node matching. We perform subgraph extraction from the target map $\mathcal{G}^t$ by selecting several candidate object nodes for each detection node in the source frame graph $\mathcal{G}^s$ based on the calculated node similarity scores. Initially, we organize vector information by creating a mapping between category labels and their respective node IDs and vectors obtained by~\eqref{eq_vec}. Specifically, the mapping employs labels as keys, each corresponding to a list of vectors. Formally, the mapping can be expressed as: 
\begin{equation}
    c \rightarrow \{(\text{ID}_i, \vec{v}_i) \mid \text{ID}_i \in \mathbb{Z}, \vec{v}_i \in \mathbb{R}^m \}.
\end{equation}
In this structure, each category $c$ corresponds to multiple nodes, each with a unique node ID and an associated graph kernel vector. This mapping allows our method to efficiently index and retrieve all nodes and their features from $\mathcal{G}^t$ according to a specific label belonging to each node of $\mathcal{G}^s$. 

For subgraph extraction, we construct a similarity matrix within each label category. Each element in the matrix represents the similarity score between detection and object nodes, determined using the cosine similarity of their respective graph kernel vectors. For each detection node $n^s_j$, a candidate set $\mathcal{N}_j^{\text{candi}}$ is generated, which encompasses these top $J$ matches and can be expressed as:
\begin{equation}
\label{eq_matches}
    \mathcal{N}_j^{\text{candi}} = \{ n^t_i \mid \text{rank}(s(n^s_j, n^t_i)) \leq J \}.
\end{equation}
Here, the similarity score $s(n^s_j, n^t_i)$ is calculated by~\eqref{eq_sim} and $\text{rank}(\cdot)$ ranks the object node $o^t_i$ among all object nodes based on similarity scores. This approach efficiently extracts subgraphs from the target map by pairing detection nodes with their most pertinent object counterparts.

\subsection{Probabilistic Node Matching and Camera Pose Refinement}
\label{subsec_refine}
Although extracting subgraphs reduces the search space, traversing all combinations like OA-SLAM~\cite{zins2022oa} remains computationally intensive. Instead, we use a RANSAC-inspired iterative approach for camera pose estimation through probabilistic node matching, involving two nested iterations: an outer iteration for random node set selection and an inner iteration for node matching and verification. The algorithm is presented in pseudocode in ~\cref{algo_ref}.

In the outer iteration, a random sampling process selects a predefined minimum number of nodes (e.g., $3$) from all detections $\mathcal{D}$. This allows the exploration of various node combinations without exhaustive searching. In the inner iteration, each selected detection node $n^s_j$ is paired with an object node $n^t_i$ randomly chosen from $\mathcal{N}_j^{\text{candi}}$, while respecting node connectivity constraints to ensure topologically plausible configurations and avoiding repeat associations of a candidate node with multiple detection nodes. A camera pose corresponding to each match set is estimated by minimizing the reprojection errors between 2D detection bounding box centers and 3D object centroids. The algorithm then updates the entire match set with all detections and counts the inliers, which are defined as matches with reprojection errors below a threshold. After completing the maximum outer and inner iterations, the best match set $\mathcal{A} = \{(n_j^s, n_i^t)\}$ and the corresponding initial camera pose $\mathbf{T}_0$ are determined based on the highest inlier count.

In the aforementioned iterative process, we use the Perspective-n-Point~(PnP) method with both 2D and 3D center points to accelerate the iterations. Finally, to achieve a more accurate camera pose, we minimize the reprojection error based on dual quadrics:
\begin{equation}
\label{eq_min_t2}
\hat{\mathbf{T}} = \mathop{\arg\min}\limits_{\mathbf{T}}\sum_{(n_j^s, n_i^t)\in\mathcal{A}} P(c_j | n_i^t) \cdot r(n_j^s, n_i^t)^2\ ,
\end{equation}
where the $c_j$ represents the category of the $j$-th detection and $P(c_j | n_i^t)$ is calculated using~\cref{eqa_prob_multi} to weigh the residual $r(n_j^s, n_i^t)^2$, which is determined by the Wasserstein distance between $\text{ell}_j^s$ from image frame and $\vec{q}_i$ from the target map considering the transformation $\mathbf{T}$, as introduced in~\cite{wang2024icra}.

\begin{algorithm}[!t]
\caption{Node Matching and Camera Pose Refinement}
\label{algo_ref}
\begin{algorithmic}[1]
\State \textbf{Input:} Detection set $\mathcal{D}=\{d_j\}_{j=1}^m$ and candidate object set $\mathcal{N}_j^{\text{candi}}$ corresponding to each detection $d_j$.

\State \textbf{Output:} Refined node matches $\mathcal{A}$ and transformation $\hat{\mathbf{T}}$
\State \textbf{Set:} Sampling detection nodes number $num$, the maximum iterations $maxIter$
\State \textbf{Intialize:} $maxInliers \leftarrow 0$, $\mathcal{A}\leftarrow \emptyset$

\While{$iter < maxIter$}
\State $\mathcal{N}^\text{tmp}\leftarrow SelectRandomNodes(\mathcal{D}, num)$ 
\ForEach{$\mathcal{A}^\text{tmp}$ in $FindCombines(\mathcal{N}^\text{tmp}, \mathcal{N}_j^{\text{candi}})$}
\State $\hat{\mathbf{T}}^\text{tmp} \leftarrow EstimatePose(\mathcal{A}^\text{tmp})$
\State $nInliers, \mathcal{A}^\text{curr} \leftarrow UpdateMatch(\mathcal{N}_j^{\text{candi}}, \hat{\mathbf{T}}^\text{tmp})$
  \If{$nInliers > maxInliers$}
  \State $\mathcal{A} \leftarrow \mathcal{A}^\text{curr}$, $maxInliers \leftarrow nInliers$
  \EndIf
\EndForEach
\EndWhile
\State \textbf{Return:} $\hat{\mathbf{T}} \leftarrow OptimizePose(\mathcal{A})$ using \eqref{eq_min_t2}
\end{algorithmic}
\end{algorithm}

\section{EXPERIMENTAL RESULTS}
\label{sec_exp}
\subsection{Experimental Setup}
\label{sec_setup}
\subsubsection{Datasets}
We conducted an extensive evaluation of our algorithm using two distinct datasets: the TUM RGB-D dataset~\cite{sturm12iros} and the Diamond dataset from ICL-Data~\cite{Characterizing19}. The TUM RGB-D dataset features sequences from real-world indoor environments. We focus on the `Fr2\_desk' and `Fr2\_person' sequences because both scenes are similar yet display slight shifts in object placement. Additionally, the `Fr2\_person' sequence includes moving people and objects, providing dynamic elements to the environment. In this experiment, we constructed an object-level map of $26$ distinct objects, encompassing over ten categories using `Fr2\_desk'. The ICL-Data Diamond dataset encompasses a variety of indoor sequences, all captured in a single scenario, but from diverse perspectives with different platforms. This includes sequences such as the handheld `Walk', the overhead `Head', the VR-based `VR', the drone-captured `DJI', and grounded `Ground'. We built an object-level map for this scenario comprising $47$ objects across $23$ different object categories and executed relocalization across various sequences using sequence `Walk'. It is worth noting that we conducted data association experiments on all sequences include `Fr2\_desk' and `Walk', but performed pose estimation experiments only on the other test sequences. Additionally, we use all frames in such sequences to evaluate the relocalization performance.

\subsubsection{Baseline Methods}
In our experiments, we evaluate the performance of GOReloc and its variants through a comprehensive comparison with several baseline methods. First, the object-level baseline methods that provide both object association and camera pose estimation include:
\begin{itemize}
    \item None Graph: The non-graph-based method. All objects sharing the same label as the detection are identified as candidates, similar to the approach presented in OA-SLAM~\cite{zins2022oa}. Here, we add our RANSAC-inspired pose refinement method for relocalization to replace the original optimization approach in OA-SLAM due to the issues encountered in handling the complex Diamond scenes. 
    \item Random Walk: The relocalization approach based on the random walk descriptors~\cite{ yu2022semanticloop, wu2023object} and the pose refinement. The random walk-based graph is used to identify candidates, followed by our proposed refinement process. The random walk's step count is set to $5$. 
    \item GOReloc: Our proposed method.
\end{itemize}
To ensure a fair comparison, both Random Walk and GOReloc employ the same subgraph extraction and pose refinement strategies, with the parameter $J$ in~\eqref{eq_matches} set to $5$. 

Furthermore, we introduce two variants of our method to validate our specific module designs:
\begin{itemize}
    \item GOReloc-NR: Our method without refinement~(No Refinement). 
    \item GOReloc-NU: Our method without considering semantic uncertainty and consistency~(No Uncertainty). 
\end{itemize}

To further explore the potential of the GOReloc application in relocalization for visual SLAM systems, we integrated it into a traditional feature-based approach. Specifically, we used GOReloc's object association results to retrieve several keyframes in the map according to the object-level co-visibility graphs proposed in VOOM~\cite{wang2024icra}. A more accurate camera pose can thus be estimated through feature matching between the source frame and these keyframes. We then compared its pose estimation performance with the traditional feature-based relocalization method utilizing the bag-of-words model used in ORB-SLAM2~\cite{mur2017orb}.

\subsection{Overall Performance of Data Association}
\label{sec_perf_da}
In this section, we focus on evaluating the performance of various multi-object data association methods. Inspired by the rigorous standards adopted in the field of multiple object tracking~(MOT)~\cite{bewley2016simple}, we employ a set of evaluation metrics specifically for object-level data association: Accuracy~(in $\%$, $\uparrow$), Center Distance~(CD, in pixels, $\downarrow$), Intersection over Union~(IoU $\uparrow$). In addition, we employ the ground-truth transformation to project the associated objects onto the image to compare them with the detections.

\begin{table}[!t]
\renewcommand{\arraystretch}{1.0}
\setlength{\tabcolsep}{5.5pt}
	\centering
	\caption{Multi-Object Data Association Performance.} 
        \vspace{-0.1cm}
	\begin{tabular}{lcccc} \toprule  \textbf{Sequence} & \textbf{Method}  & \makecell[c]{\textbf{Accuracy}($\%$)\\ ($\uparrow$)}  &  \makecell[c]{\textbf{CD}(pixels)\\($\downarrow$)}   & \makecell[c]{\textbf{IoU}  \\ ($\uparrow$) }  
 \\
		\midrule
		 \multirow{5}{*}{Walk}  & None Graph &  33.59 & 261.09 & 0.106 \\
        & Random Walk  &  26.79 & 325.38 &   0.056 \\
       & GOReloc-NR &  32.54  & 196.04  &  0.094 \\ 
		 & GOReloc (Ours)   &  \textbf{43.71} & \textbf{167.35}  & \textbf{0.148}
   \\
   \midrule
		 \multirow{5}{*}{Head}  & None Graph & 26.63 & 323.78 & 0.070 \\
        & Random Walk & 20.56 & 333.20 &  0.037\\
       & GOReloc-NR &  23.40  & 258.64 &  0.060 \\
		 & GOReloc (Ours)  &  \textbf{32.51} & \textbf{208.83}  & \textbf{0.086}
   \\
   \midrule
		 \multirow{5}{*}{VR}  & None Graph  &  18.40 & 385.04 & 0.046 \\
        & Random Walk & 17.06 & 396.18 &  0.030 \\
       & GOReloc-NR &  21.37    & 268.96 &  0.056 \\
		 & GOReloc (Ours)   &  \textbf{23.16} & \textbf{288.61}  & \textbf{0.061}
   \\
   \midrule
		 \multirow{5}{*}{DJI}   & None Graph &  13.50 & 278.27 & 0.025 \\
        & Random Walk  & 10.37 & 299.11 & 0.017 \\
       & GOReloc-NR &  10.97    & 262.29 &  0.013 \\
		 & GOReloc (Ours)   &  \textbf{14.70} & \textbf{259.35}  & \textbf{0.027}
   \\
		\midrule
		 \multirow{5}{*}{Fr2\_desk} & None Graph &  55.88 & 89.99 & 0.159 \\
        & Random Walk  & 37.76 & 192.90 &  0.056 \\
       & GOReloc-NR &  40.59   & 150.80 &  0.116 \\
		 & GOReloc (Ours)  & \textbf{61.88} & \textbf{73.53}   & \textbf{0.186}
   \\ 
		\bottomrule
	\end{tabular} \\
     \begin{tablenotes} 
     \item Higher $\uparrow$ and lower $\downarrow$ indicate preferred metrics; top results of each sequence are \textbf{bolded}.  
     \end{tablenotes}
 \label{table_asso}
\vspace{-0.3cm}
\end{table}

~\cref{table_asso} presents the multi-object data association performance of different methods based on various metrics. Across all sequences, GOReloc consistently demonstrates superior performance in all key metrics. While GOReloc delivers the best results, GOReloc-NR, which operates without refinement steps, often surpasses None Graph and Random Walk, both of which utilize optimization. This underscores the fundamental strength of the GOReloc approach, showing that even in its unrefined state, the core algorithm is robust enough to outperform traditional methods that require more complex processing or refinement.

In contrast, traditional methods like None Graph and Random Walk consistently show lower performance across the board. For example, in the `VR' sequence, Random Walk records the worst metrics of all methods, demonstrating the limitations of the traditional graph-based data association techniques in more dynamic or cluttered environments. Interestingly, the None Graph method outperforms Random Walk in many metrics. This is because None Graph has accepted all objects with the same label as candidates, unlike graph-based methods like Random Walk and our methods, which select only the top $5$ most similar objects as candidates for each detection. Despite this, GOReloc still outperforms None Graph, underscoring the effectiveness of our graph-based method in selecting precise association candidates. 

\begin{figure*}[!t]
    \centering
    \begin{minipage}{.015\textwidth}
        \centering
        \vspace{0.5mm}
        \rotatebox{90}{\textbf{TUM RGB-D}}
    \end{minipage} \noindent \hspace{-1mm}
    \begin{minipage}{.002\textwidth}
        \centering
        \vspace{0.5mm}
        \rule{0.4pt}{6.2cm}
    \end{minipage} \noindent \hspace{-3mm}
    \begin{minipage}{.96\textwidth}
        \centering
        \begin{minipage}{.32\textwidth}
            \centering
            \textbf{None Graph}
        \end{minipage} \noindent  \hspace{0.5mm}
        \begin{minipage}{.32\textwidth}
            \centering
            \textbf{Random Walk}
        \end{minipage} \noindent  \hspace{0.5mm}
        \begin{minipage}{.32\textwidth}
            \centering
            \textbf{GOReloc (Ours)}
        \end{minipage} \\ \vspace{1.2mm}
        \subfloat{%
      \includegraphics[width=0.32\linewidth]{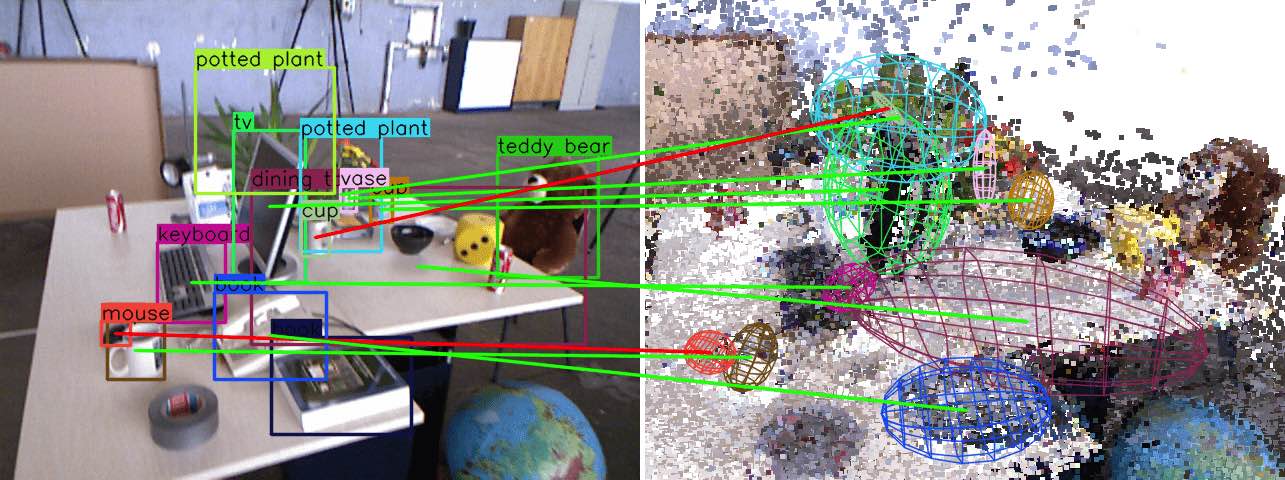}}\noindent  \hspace{0.5mm}
      \subfloat{%
        \includegraphics[width=0.32\linewidth]{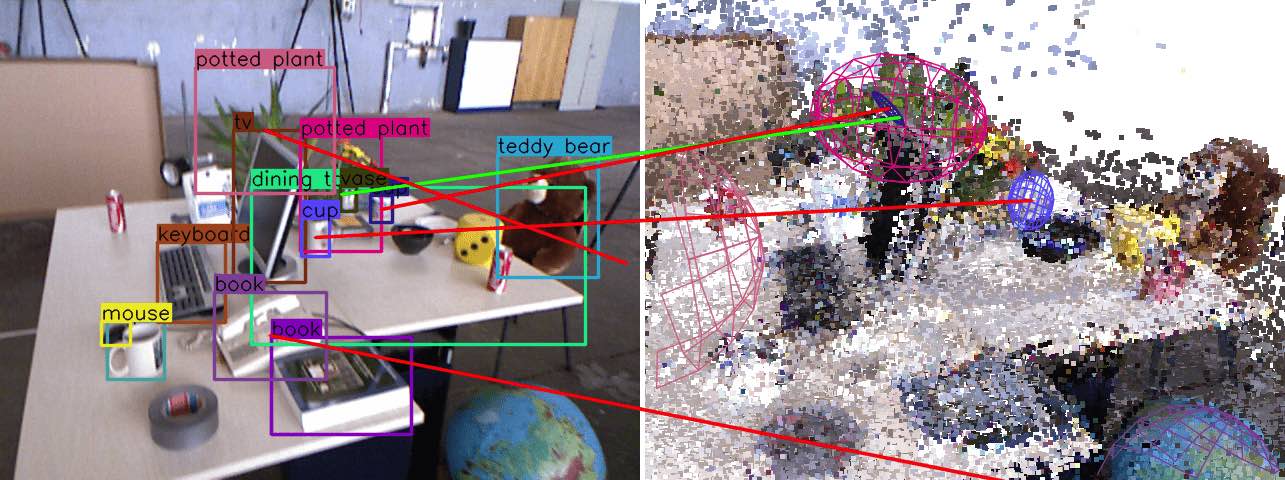}}\noindent  \hspace{0.5mm}
      \subfloat{%
      \includegraphics[width=0.32\linewidth]{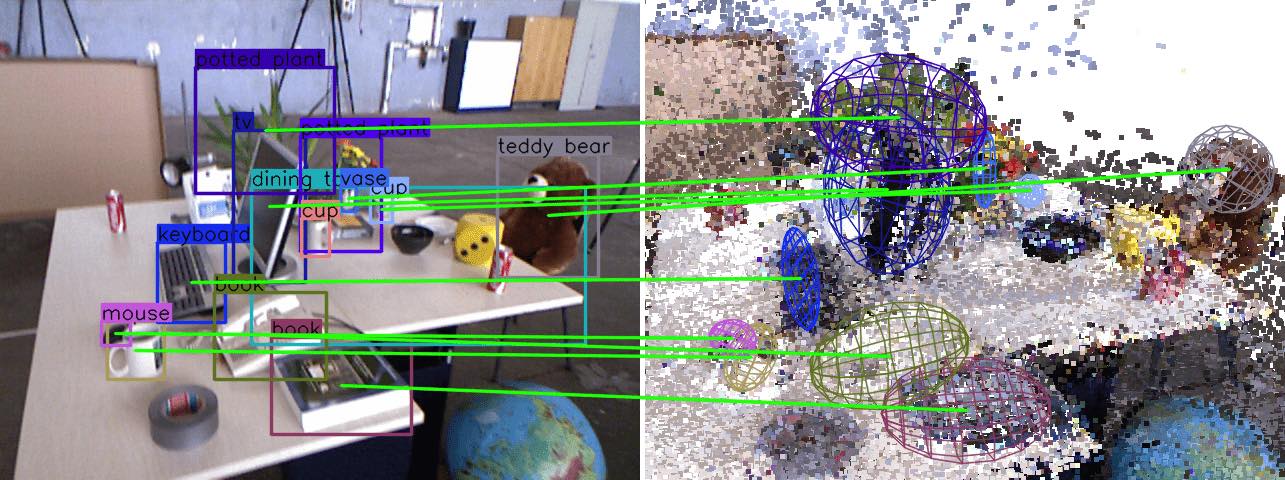}}
      \\ \vspace{0.5mm}
      \subfloat{%
      \includegraphics[width=0.32\linewidth]{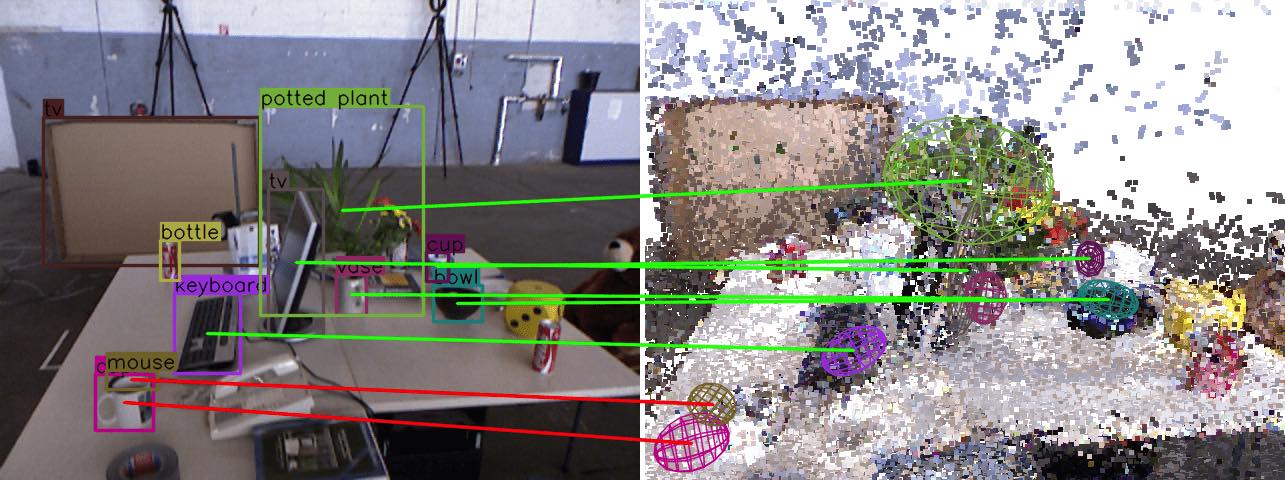}}\noindent  \hspace{0.5mm}
      \subfloat{%
        \includegraphics[width=0.32\linewidth]{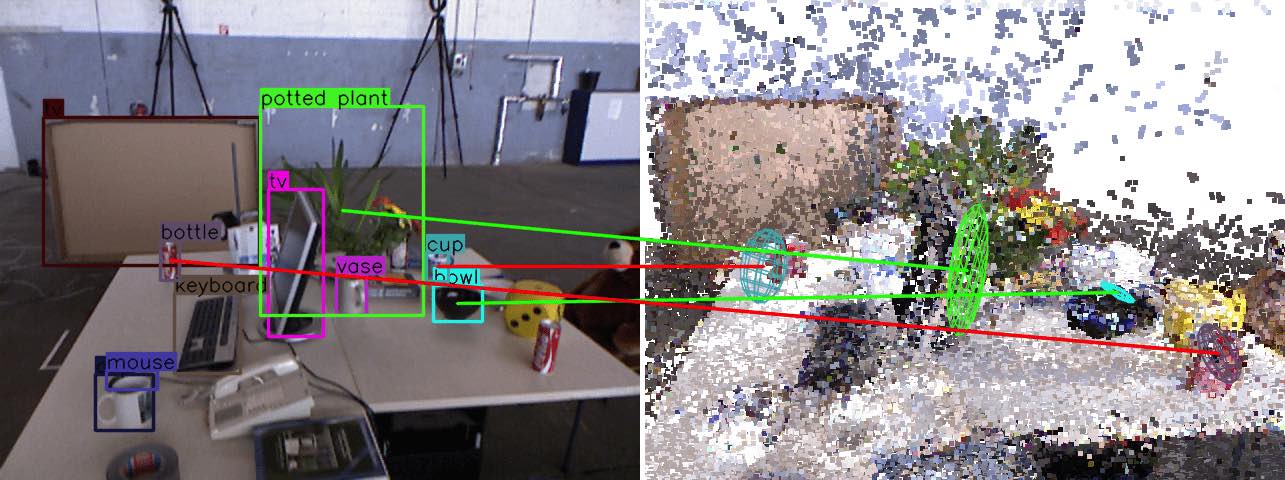}}\noindent  \hspace{0.5mm}
      \subfloat{%
      \includegraphics[width=0.32\linewidth]{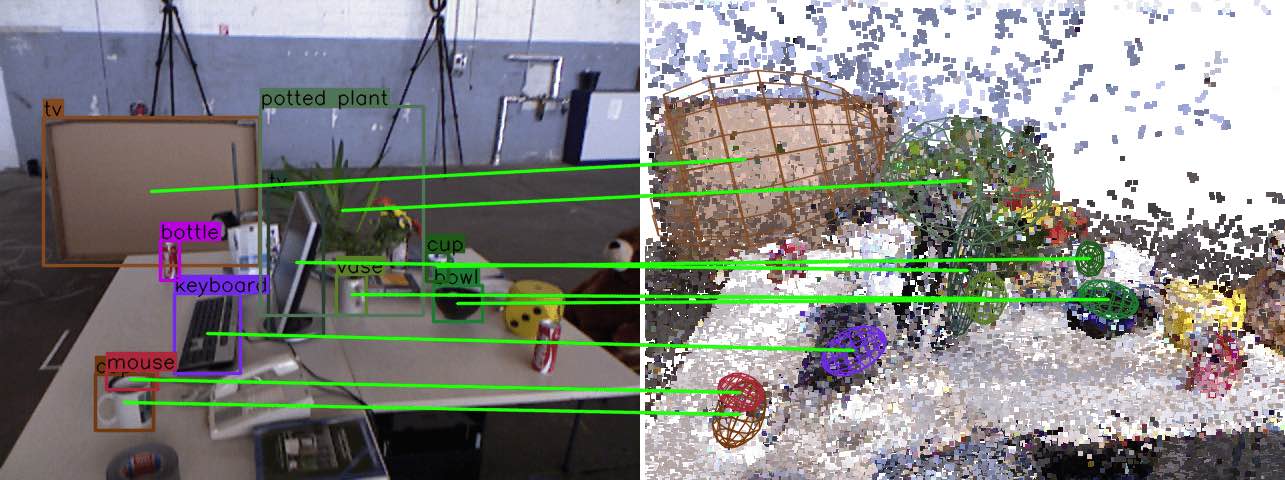}}
      \\ \vspace{0.5mm}
      \subfloat{%
      \includegraphics[width=0.32\linewidth]{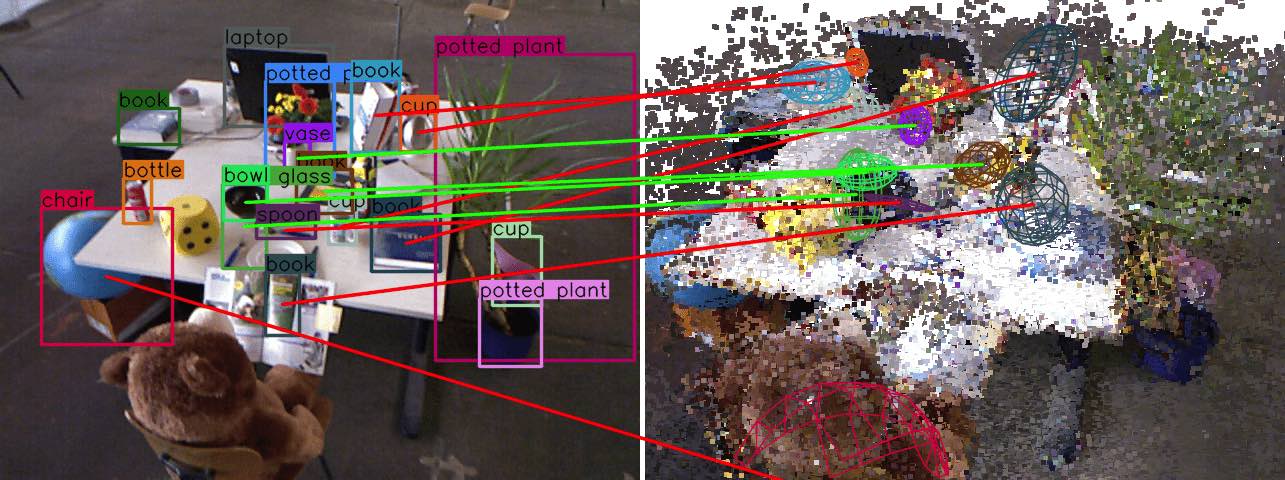}}\noindent  \hspace{0.5mm}
      \subfloat{%
        \includegraphics[width=0.32\linewidth]{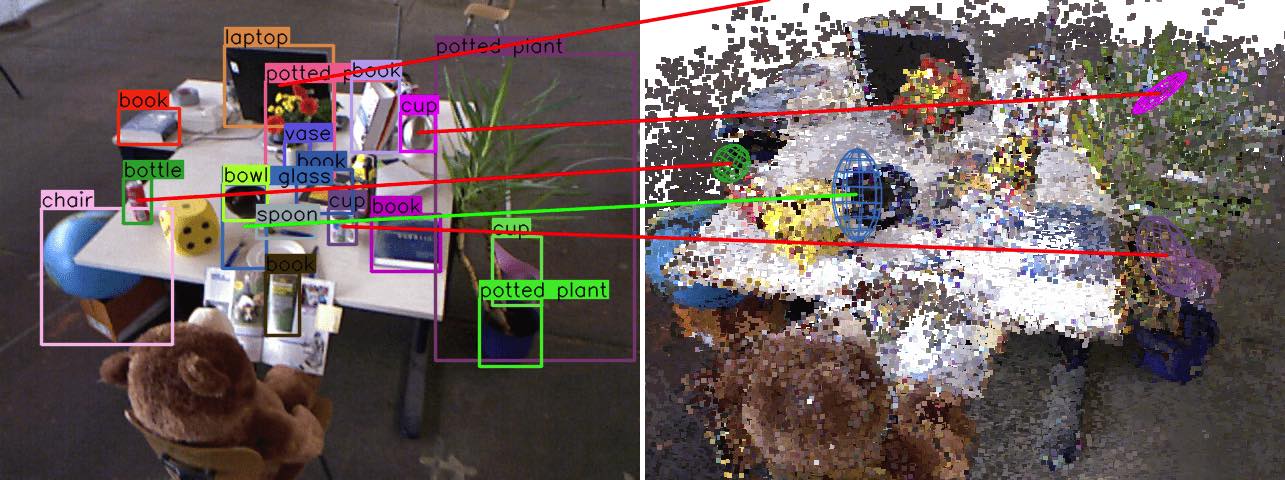}}\noindent  \hspace{0.5mm}
      \subfloat{%
      \includegraphics[width=0.32\linewidth]{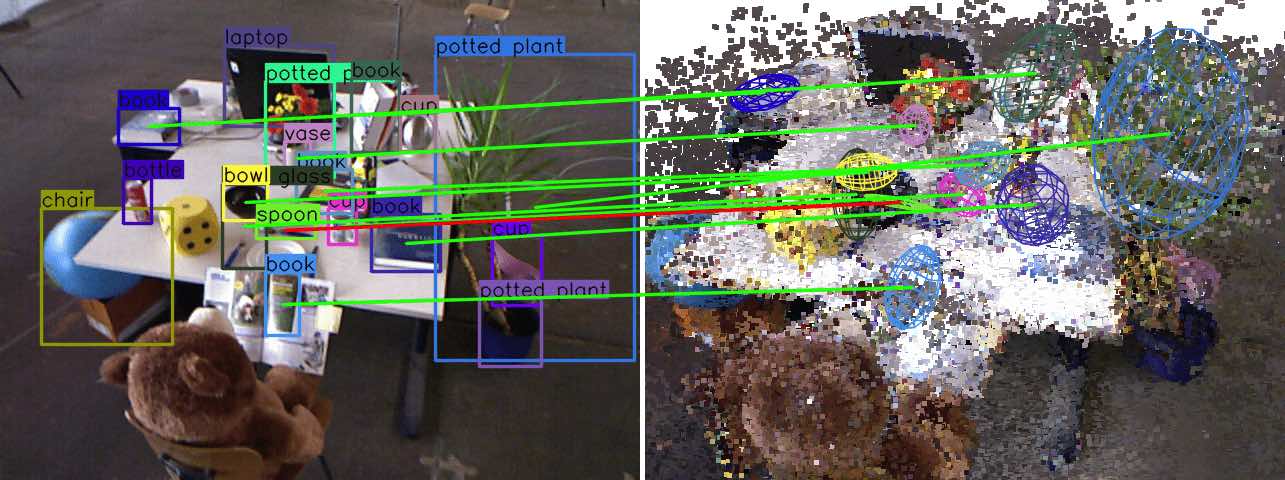}}
      \\ \vspace{0.5mm}
    \end{minipage}
\\
    \begin{minipage}{.015\textwidth}
        \centering
        \vspace{0.5mm}
        \rotatebox{90}{\textbf{ICL-Data}}
    \end{minipage} \noindent \hspace{-1mm}
    \begin{minipage}{.002\textwidth}
        \centering
        \vspace{0.5mm}
        \rule{0.4pt}{6.2cm}
    \end{minipage} \noindent \hspace{-3mm}
    \begin{minipage}{.96\textwidth}
        \centering
        \subfloat{%
      \includegraphics[width=0.32\linewidth]{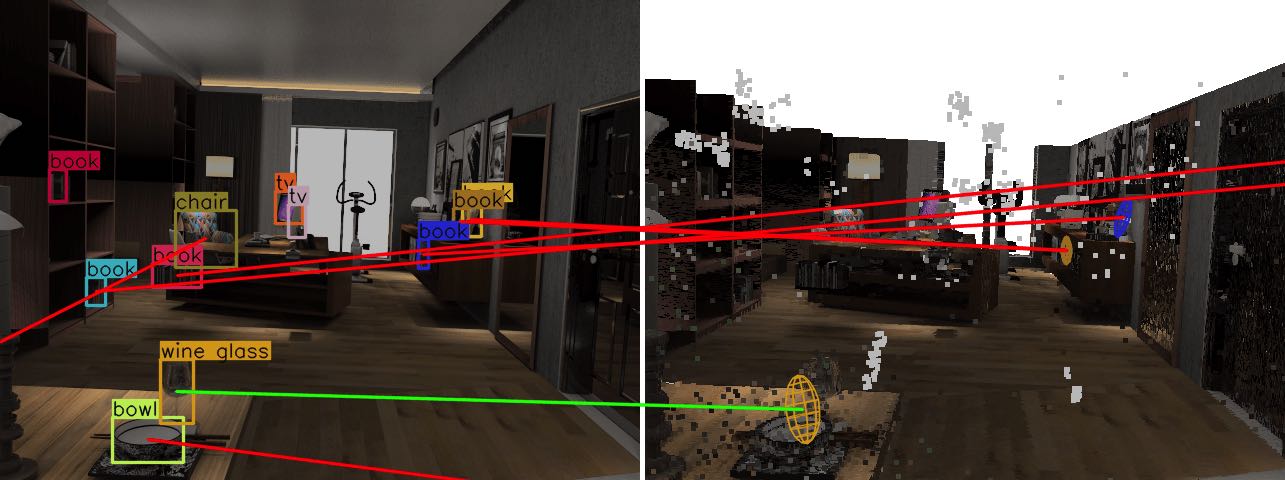}}\noindent  \hspace{0.5mm}
      \subfloat{%
        \includegraphics[width=0.32\linewidth]{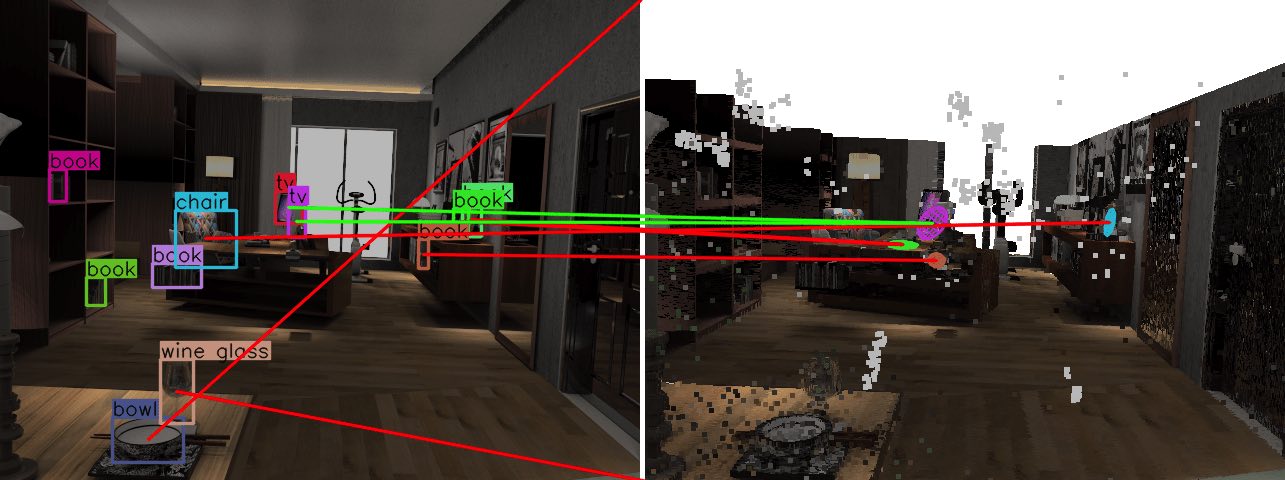}}\noindent  \hspace{0.5mm}
      \subfloat{%
      \includegraphics[width=0.32\linewidth]{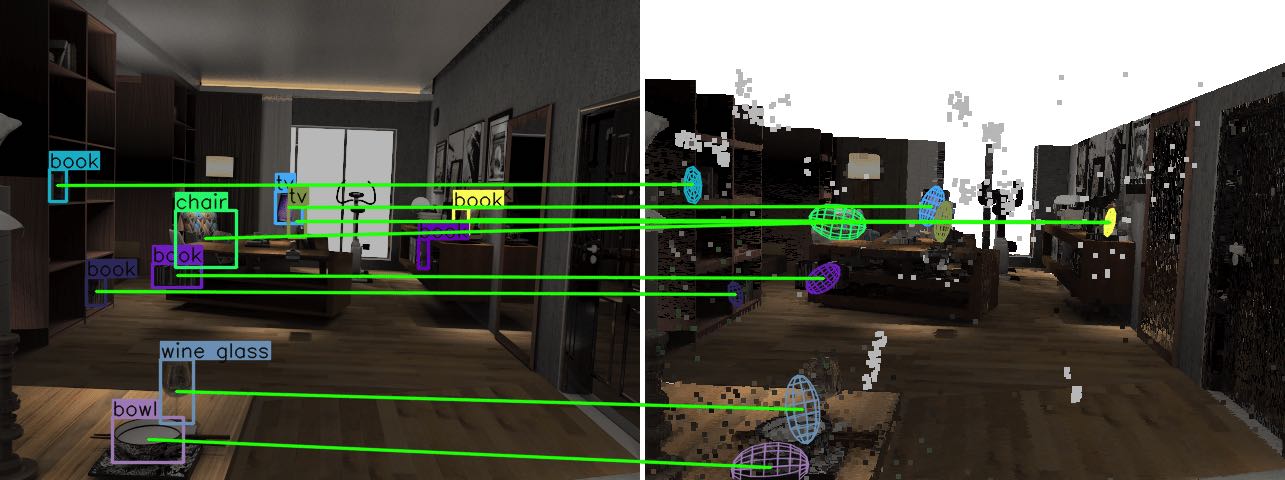}}
      \\ \vspace{0.5mm}
        \subfloat{%
      \includegraphics[width=0.32\linewidth]{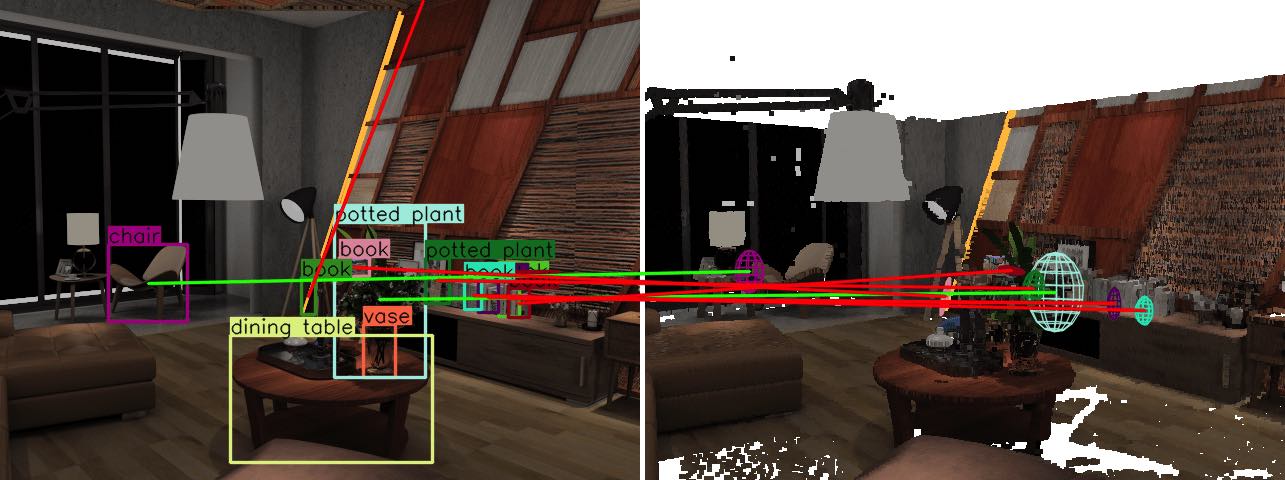}}\noindent  \hspace{0.5mm}
      \subfloat{%
        \includegraphics[width=0.32\linewidth]{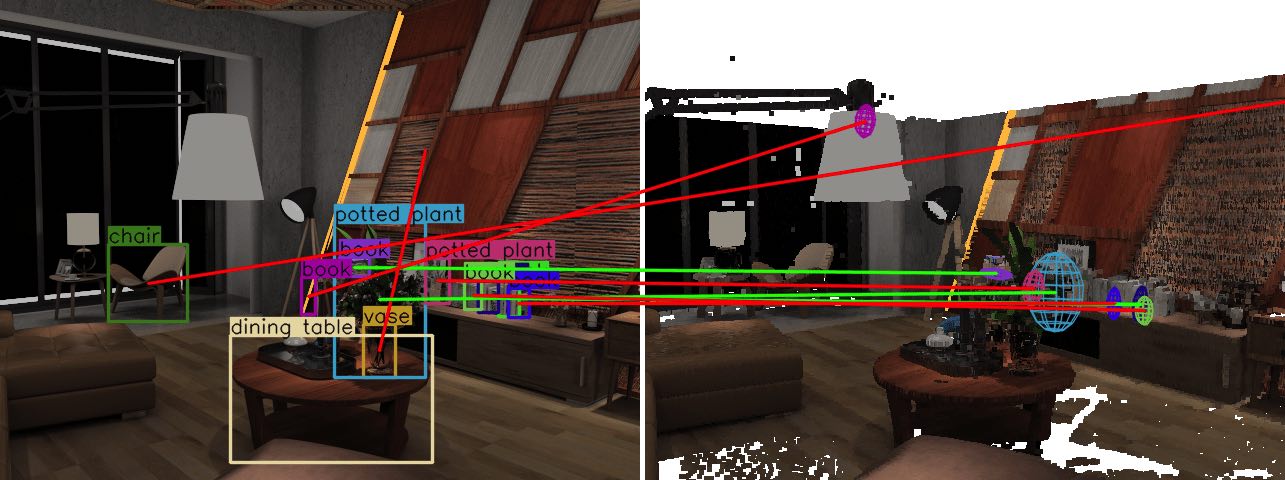}}\noindent  \hspace{0.5mm}
      \subfloat{%
      \includegraphics[width=0.32\linewidth]{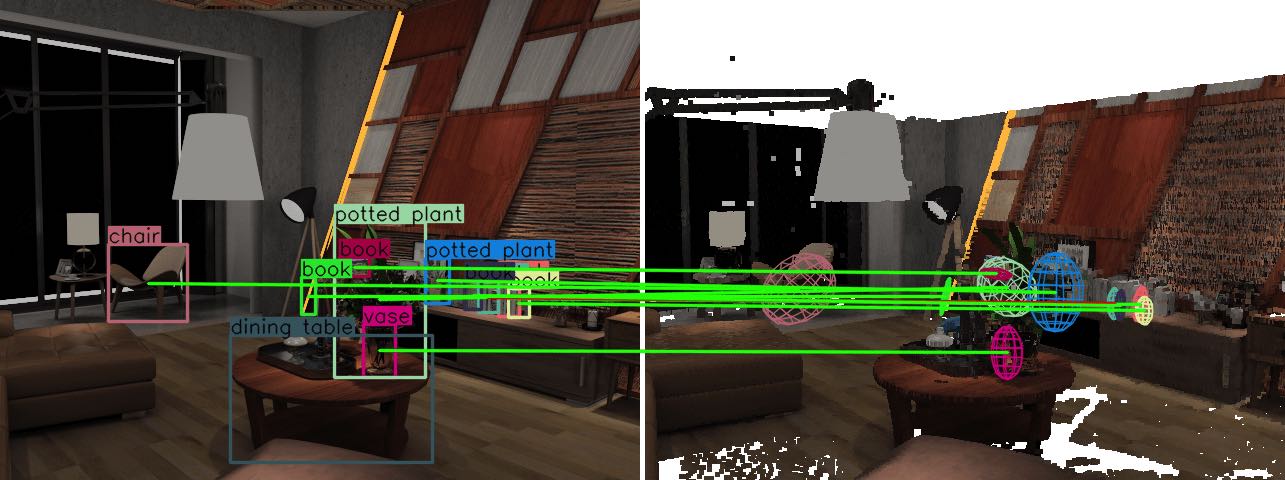}}
      \\ \vspace{0.5mm}
      \subfloat{%
      \includegraphics[width=0.32\linewidth]{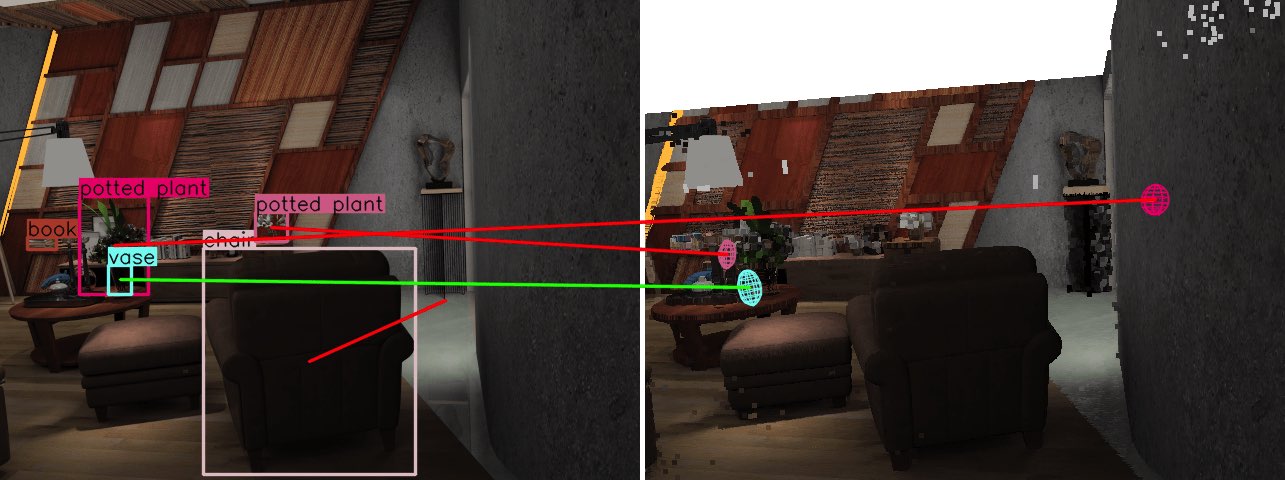}}\noindent  \hspace{0.5mm}
      \subfloat{%
        \includegraphics[width=0.32\linewidth]{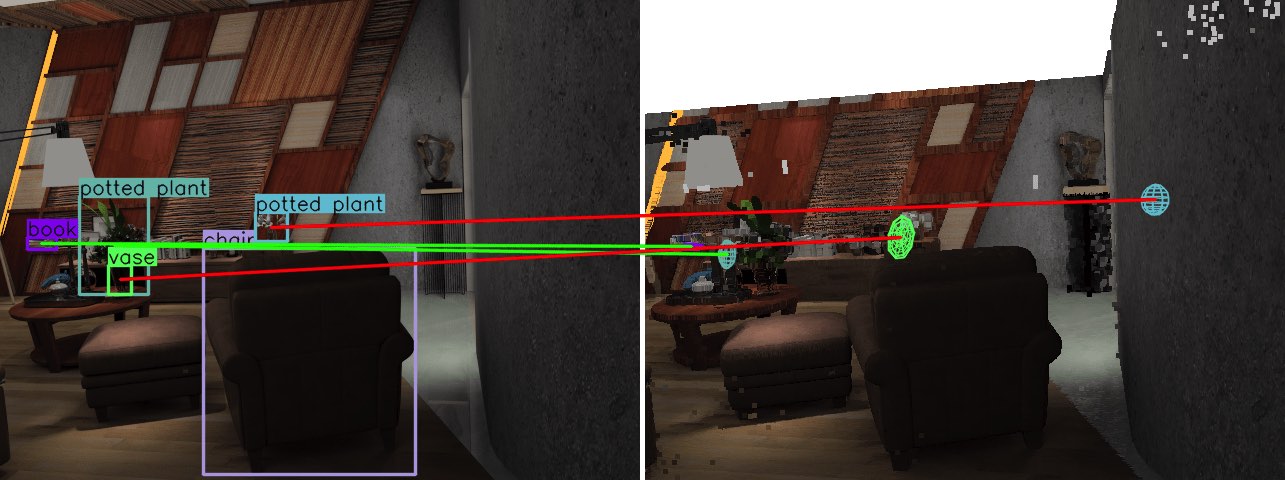}}\noindent  \hspace{0.5mm}
      \subfloat{%
      \includegraphics[width=0.32\linewidth]{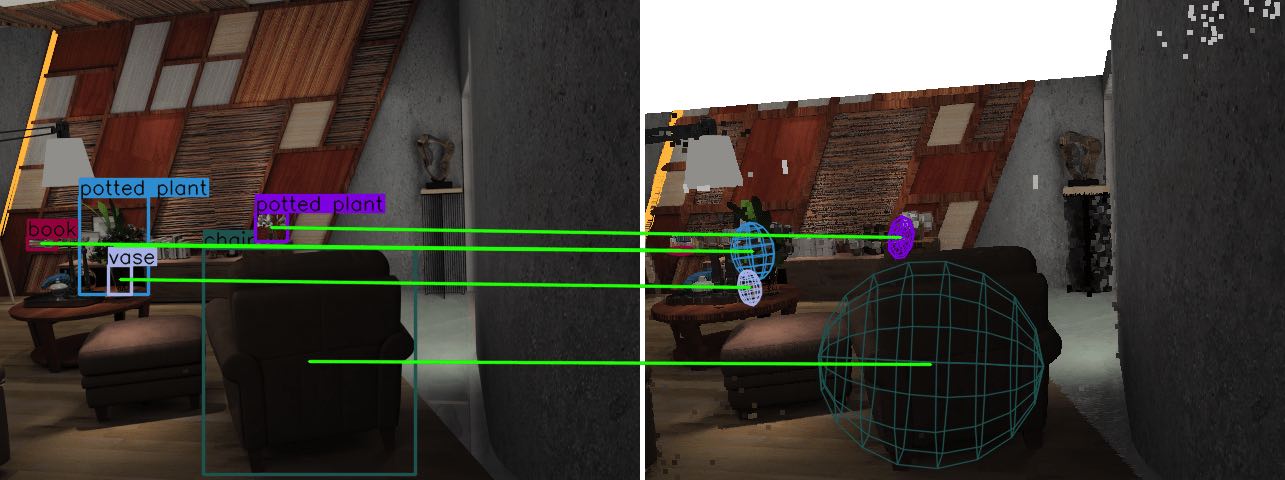}}
    \end{minipage}
    \caption{The qualitative results of object-level data association. In each pair, the left is the frame image, while the right is the map rendering. Correct associations are marked with green lines, and incorrect associations are marked with red lines.}
    \label{fig:quali}
    \vspace{-0.3cm}
\end{figure*}

~\cref{fig:quali} showcases the qualitative outcomes of object-level data association compared to None Graph and Random Walk methods, featuring multiple matching pairs from both datasets. In each pair, the frame image is displayed on the left. The corresponding rendering image on the right is created by projecting the point cloud~(only used for visualization) and objects using the ground-truth transformation matrices. Associations are visually annotated: correct results are marked with green lines and incorrect associations with red lines. Intuitively, our method demonstrates a higher overall accuracy rate in these scenarios. 

\subsection{Overall Performance of Pose Estimation}
\label{sec_perf_pe}

This section examines the overall performance of pose estimation. First, we compare the GOReloc method with other object-level approaches. Subsequently, we evaluate GOReloc performance within a visual SLAM system and compare the results with those of ORB-SLAM2's relocalization.

\subsubsection{Comparison of Pose Estimation with Object-level Methods}
\begin{table}[!t]
\renewcommand{\arraystretch}{1.0}
\setlength{\tabcolsep}{6.5pt}
	\centering
	\caption{Comparison of Pose Estimation with Object-level Methods} 
        \vspace{-0.1cm}
	\begin{tabular}{cccccc} \toprule \multirow{2}{*}{\textbf{Sequence}} & \multirow{2}{*}{\textbf{Method}} & \multicolumn{2}{c}{\textbf{Success Rate} ($\%$) $\uparrow$} & \multicolumn{2}{c}{\textbf{TE}(\SI{}{\meter}) $\downarrow$}  
 \\ 
 \cmidrule(lr){3-4}\cmidrule(lr){5-6} 
    & & @2m & @5m & $10\%$ & $20\%$ 
 \\ 
  \midrule
  \multirow{3}{*}{Head} & None Graph & 11.31 & 27.98 & 1.73 & 3.27\\
        & Random Walk & 5.35 & 18.65 & 3.34 & 5.08 \\
	& GOReloc-NU   & 16.67 & 36.54 & \textbf{1.05} & 2.39 \\
	& GOReloc~(Ours)   & \textbf{17.43} & \textbf{38.23} & 1.12 & \textbf{2.28} 
   \\ 
   \midrule
   \multirow{3}{*}{VR} & None Graph & 4.98 & 25.65 & 2.99 & 4.33  \\
   & Random Walk & 1.75 & 20.67 & 3.71  & 4.88 \\
   & GOReloc-NU   & 8.90 & 30.97 & 2.13  & 3.38  \\
	 & GOReloc~(Ours) & \textbf{9.46} & \textbf{36.37} & \textbf{1.96}  & \textbf{3.30} 
   \\ \midrule
 \multirow{3}{*}{DJI} & None Graph & 3.81 & 13.75 & 3.03 & 4.48\\
        & Random Walk & 2.98 & 20.13 & 3.30 & 4.97 \\
        & GOReloc-NU   & 7.24 & 31.05  & 2.33 & 3.61  \\
	& GOReloc~(Ours)   & \textbf{9.65} & \textbf{37.33}  & \textbf{2.02} & \textbf{2.99} 
   \\   
   \midrule
 \multirow{3}{*}{Ground} & None Graph & 4.58 & 25.78 & 3.04 & 4.45\\
        & Random Walk & 1.93 & 27.23 & 3.59 & 4.47 \\
        & GOReloc-NU  & 6.51 & 26.75  & 2.42 & 4.33  \\
	& GOReloc~(Ours)   & \textbf{7.95} & \textbf{32.29}  & \textbf{2.16} & \textbf{3.06} 
   \\  
		\midrule
 \multirow{3}{*}{Fr2\_person} & None Graph & 50.42 & 95.42 & 0.82 & 1.04  \\
     & Random Walk & 20.22 & 83.72 & 1.34  & 1.99  \\
     & GOReloc-NU   & 54.75 & \textbf{96.19} & 0.76  & 0.97  \\
	& GOReloc~(Ours)  & \textbf{64.87} & 96.11 & \textbf{0.73}  & \textbf{0.90} 
   \\ 
		\bottomrule
	\end{tabular} \\
     \begin{tablenotes} 
     \item Higher $\uparrow$ and lower $\downarrow$ indicate preferred metrics; The best results are \textbf{bolded}.  
     \end{tablenotes}
 \label{table_pose}
\vspace{-0.3cm}
\end{table}

Our comparison of object-level approaches on pose estimations is shown in ~\cref{table_pose}. The metrics include the Success Rate and Translation Error~(TE)~\cite{xu2023ring}. Success rates are measured under conditions where TE is less than $2$ meters (@$2$m) and $5$ meters (@$5$m), indicating the frequency of accurate pose estimations within these limits. The TE columns labeled 10\% and 20\% indicate the average translation errors within the best 10\% and 20\% of results, respectively. 

The table clearly shows that GOReloc and GOReloc-NU consistently exhibit superior performance across the tested sequences. Notably, they achieve the top two success rates in ICL-Data, particularly excelling at the 2-meter mark, where precision is critical. It is noteworthy that the success rate on the `Fr2\_person' sequence from the TUM RGB-D dataset is significantly higher at 64.87\% for distances under 2 meters, in contrast to some ICL-Data sequences, which fall below 10\%. This difference highlights the relocalization challenges in `Diamond' scenarios with numerous objects, high dynamics (VR and DJI), and limited views (Ground). This underscores that our algorithm not only tackles simple scenarios effectively but also adapts to complex environments, despite the varying degrees of performance. In terms of TE, both versions of GOReloc consistently outperform the other methods, maintaining lower errors that underscore their precision. GOReloc often achieves the lowest TE across the evaluated sequences, showcasing its reliability in providing accurate pose estimations. 

While both GOReloc versions outperform traditional methods, GOReloc generally achieves better results than GOReloc-NU, demonstrating that incorporating semantic uncertainty and consistency improves accuracy and reliability in complex environments. This is most apparent in scenarios like the `Ground' sequence, where the limited perspectives of ground robots lead to less reliable detections. However, in situations with more reliable detections, such as the `Head' sequence where GOReloc-NU outperforms GOReloc by $0.07$ at the 10\% threshold, the addition of semantic uncertainty and consistency might be less advantageous.

\subsubsection{Comparison of Pose Estimation with Feature-based Methods}

\begin{table}[!ht]
\renewcommand{\arraystretch}{1.0}
\setlength{\tabcolsep}{7pt}
	\centering
	\caption{Comparison of Pose Estimation with Feature-based Method} 
        \vspace{-0.1cm}
	\begin{tabular}{cccccc} \toprule \multirow{2}{*}{\textbf{Sequence}} & \multirow{2}{*}{\textbf{Method}} & \multicolumn{4}{c}{\textbf{Success Rate} ($\%$) $\uparrow$}
 \\ 
 \cmidrule(lr){3-6} 
    & & @0.5m & @1m & @2m & @3m 
 \\ 
		
  \midrule
  \multirow{2}{*}{Head} & ORB-SLAM2 & 48.05 & 48.82 & 48.89 & 48.89
  \\
	& GOReloc~(Ours) & \textbf{52.33} & \textbf{54.62} &\textbf{55.77} & \textbf{56.76}
   \\ 
   \midrule
   \multirow{2}{*}{VR} & ORB-SLAM2 & 28.71 & 28.95 & 28.95 & 28.95 
  \\
	& GOReloc~(Ours) & \textbf{33.56} & \textbf{35.33} & \textbf{36.73} & \textbf{39.41} 
 \\ 
   \midrule
   \multirow{2}{*}{DJI} & ORB-SLAM2 & 6.10 & 6.10 & 6.10 & 6.10 
  \\
	& GOReloc~(Ours) & \textbf{7.70} & \textbf{8.95} & \textbf{11.99} & \textbf{16.04} 
   \\ 
   \midrule
 \multirow{2}{*}{Ground} & ORB-SLAM2 & 2.44 & 2.69 & 2.69 & 2.69
  \\
	& GOReloc~(Ours) & \textbf{3.13} & \textbf{4.25} & \textbf{5.06} & \textbf{6.88}
   \\   
		\midrule
 \multirow{2}{*}{Fr2\_person} & ORB-SLAM2 & 1.97 & 6.42 & 9.12 & 9.17 
  \\
	& GOReloc~(Ours) & \textbf{3.52} & \textbf{22.79} & \textbf{53.26} & \textbf{70.17} 
   \\ 
		\bottomrule
	\end{tabular} \\
     \begin{tablenotes} 
     \item Higher $\uparrow$ indicate preferred metrics; The best results are \textbf{bolded}.  
     \end{tablenotes}
 \label{table_pose_f}
\vspace{-1em}
\end{table}

~\cref{table_pose_f} shows a comparison between our GOReloc and the feature-based ORB-SLAM2 across multiple sequences, evaluating success rates at distance thresholds of $0.5$m, $1$m, $2$m, and $3$m. Across all tested scenarios, GOReloc consistently outperforms ORB-SLAM2, demonstrating significant improvement of the proposed object-aided approach. For instance, in the `Fr2\_person' sequence, GOReloc's success rate at $3$m remarkably increases to $70.17\%$ compared to ORB-SLAM2's $9.17\%$. The results demonstrate GOReloc's potential in challenging dynamic environments, particularly where traditional methods fail. It should also be noted that integrating GOReloc with feature-based visual SLAM achieves higher success rates than using only objects, demonstrating that objects boost robustness while features enhance accuracy, highlighting their complementary roles in relocalization.

\subsection{Runtime Analysis}
\label{sec_time_ana}
We tested our method using a laptop with an Intel i9-10885H CPU. Details for the runtime analysis are presented in~\cref{table_time}. GOReloc maintains a high processing rate at 52.6\,Hz for TUM RGB-D `Fr2\_desk' and 57\,Hz for `Diamond Walk', showcasing its capability to function well in real-time applications. In comparison, when tested with OA-SLAM~\cite{zins2022oa} on the TUM RGB-D dataset, each relocalization process runs with a significantly lower rate of 6.2\,Hz. Moreover, even though the `Diamond' scene only has twice the number of objects as the `Fr2\_desk' scene, OA-SLAM directly timed out when running on it, failing to produce the final results. This comparison indicates the superior speed of our GOReloc compared to the OA-SLAM approach.

\begin{table}[!ht]
\renewcommand{\arraystretch}{1.0}
\setlength{\tabcolsep}{6.5pt}
\centering
\caption{Runtime Analysis} 
\begin{tabular}{lcc}
\toprule
                     & TUM RGB-D & ICL-Data \\  \midrule
Frame Processing        &   12.811\,ms     &   11.213\,ms           \\
Graph Generation     & 0.071\,ms    & 0.038\,ms       \\
Subgraph Extraction & 0.021\,ms    & 0.032\,ms       \\
Refinement         & 6.122\,ms    & 6.257\,ms       \\
\midrule
Total                & 19.025\,ms (52.6\,Hz)   & 17.540\,ms (57.0\,Hz)  \\ \bottomrule        
\end{tabular}
 \label{table_time}
\end{table}

\section{Conclusion}
\label{sec_conc}
This article proposed a graph-based association method for object-level camera relocalization systems. It introduced a novel graph-based node-matching technique considering semantic uncertainty and consistency and a pose refinement algorithm for object-level data association and camera pose estimation. Utilizing these components, a complete real-time object-level relocalization system was achieved. The experimental results illustrate that GOReloc and its variants consistently outperform conventional object-level methods, particularly in complex and dynamic environments. The superior pose estimation performance of GOReloc highlights the effectiveness of integrating object-level data association and semantic analysis into the visual SLAM process.

\bibliographystyle{IEEEtran}
\bibliography{wang2024goreloc}

\end{document}